\documentclass[runningheads]{llncs}
\usepackage{graphicx}
\usepackage{tikz}
\usepackage{comment}
\usepackage{amsmath,amssymb} % define this before the line numbering.
\usepackage{color}
\usepackage{url}
\usepackage[hidelinks]{hyperref}
\usepackage{graphicx}
\usepackage{bm}
\usepackage{booktabs}
\usepackage{multirow}
\usepackage{pifont}
\newcommand{\cmark}{\ding{51}}%

\usepackage{xspace}

\makeatletter
\DeclareRobustCommand\onedot{\futurelet\@let@token\@onedot}
\def\@onedot{\ifx\@let@token.\else.\null\fi\xspace}

\def\eg{\emph{e.g}\onedot} 
\def\ie{\emph{i.e}\onedot} 
 
\def\etc{\emph{etc}\onedot} 
 
\def\etal{\emph{et al}\onedot}
\makeatother

\newcommand{\mname}{VIKING}

\begin{document}
% \linenumbers
\pagestyle{headings}
\mainmatter
\def\ECCVSubNumber{13}  % Insert your submission number here

\title{A Dataset and Baselines for\\Visual Question Answering on Art} % Replace with your title

\begin{comment}
\titlerunning{ECCV-20 submission ID \ECCVSubNumber} 
\authorrunning{ECCV-20 submission ID \ECCVSubNumber} 
\author{Anonymous ECCV submission}
\institute{Paper ID \ECCVSubNumber}
\end{comment}
%******************

% \begin{comment}
\titlerunning{Visual Question Answerng on Art}
\author{Noa Garcia\inst{1} \and
Chentao Ye\inst{2} \and
Zihua Liu\inst{2} \and
Qingtao Hu\inst{2} \and
Mayu Otani\inst{3} \and \\
Chenhui Chu\inst{1} \and
Yuta Nakashima\inst{1} \and
Teruko Mitamura\inst{2}
}
\authorrunning{Garcia et al.}
\institute{Osaka University, Japan \and
Carnegie Mellon University, USA \and
CyberAgent, Inc., Japan}
% \end{comment}
%******************
\maketitle

\begin{abstract}
Answering questions related to art pieces (paintings) is a difficult task, as it implies the understanding of not only the visual information that is shown in the picture, but also the contextual knowledge that is acquired through the study of the history of art. In this work, we introduce our first attempt towards building a new dataset, coined AQUA (Art QUestion Answering). The question-answer (QA) pairs are automatically generated using state-of-the-art question generation methods based on paintings and comments provided in an existing art understanding dataset. The QA pairs are cleansed by crowdsourcing workers with respect to their grammatical correctness, answerability, and answers' correctness. Our dataset inherently consists of visual (painting-based) and knowledge (comment-based) questions. We also present a two-branch model as baseline, where the visual and knowledge questions are handled independently. We extensively compare our baseline model against the state-of-the-art models for question answering, and we provide a comprehensive study  about the challenges and potential future directions for visual question answering on art.

\keywords{Visual question answering, art dataset, external knowledge}
\end{abstract}

\section{Introduction}

Providing human-like semantic interpretation of visual information is one of the ultimate goals of technologies around artificial intelligence, computer vision, and natural language processing. Tremendous research efforts have been made towards this goal, including object detection \cite{PASCAL-VOC}, phrase grounding \cite{Plummer_2015_ICCV}, image/video captioning \cite{Vinyals_2015_CVPR}, \etc. Visual question answering (VQA) is among these works and is now one of the main stream topics \cite{Wu:2017:CVIU}. VQA may require high-level comprehension of the image content, as well as questions given as natural language. Recently, various extensions of the VQA task have been proposed, including ones requiring knowledge \cite{wu2016ask}. 

The main target of the VQA task has been natural images, which capture real-world objects, scenes, and events. Very few work addresses other types of visual information, \eg, the abstract image subset of the VQA dataset \cite{agrawal2015vqadataset}, CLEVR \cite{johnson2017clevr}, and PororoQA \cite{kim2017deepstory}. One of the primary reasons for using non-real-world images is to unburden visual recognition in the VQA pipeline to give more focus on answer prediction.  

Meanwhile, artworks, or paintings, are another interesting domain for VQA. Besides their cultural and historical importance, paintings pose extra challenges in VQA: Firstly, paintings may express a subject in different abstraction levels, perhaps being associated to the continuum spanned by naturalism, realism, symbolism, impressionism, cubism, \etc. Pretrained models for, \eg, object detection, may work well for realism but not necessarily for cubism. Secondly, the interpretation of paintings can be highly dependent on their background, such as the social and the author's personal context, which may not be fully conveyed from the paintings themselves. This implies that external knowledge on the background of paintings may be needed for answering questions. 

\begin{figure}[t]
    \centering
    \includegraphics[width=0.85\columnwidth]{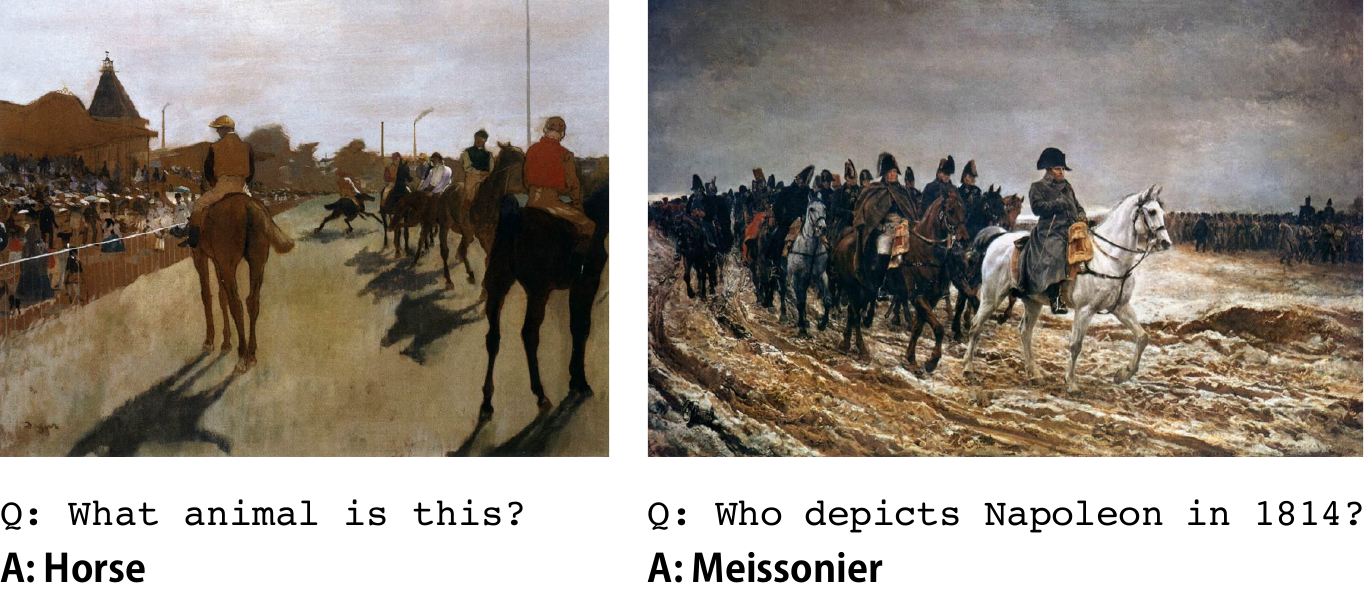}
    \caption{Examples from the AQUA dataset. There are two different types of QA pairs: generated from paintings (left) and generated from paintings' comments (right). }
    \label{fig:example_qa_pairs}
\end{figure}

This paper offers our first attempt to build a benchmark for question answering on artworks by providing the AQUA (\textbf{a}rt \textbf{qu}estion \textbf{a}nswering) dataset, built upon the SemArt dataset \cite{Garcia2018How}, together with a baseline model. The QA pairs (see examples in Figure \ref{fig:example_qa_pairs}) are automatically generated using multiple state-of-the-art question generation methods, which have been studied in the communities of both natural language processing \cite{heilman-smith-2010-good,du-etal-2017-learning,zhao-etal-2018-paragraph,lewis2019generative} 
and computer vision \cite{johnson2017clevr,yang2018curiosity,misra2018learning,krishna2019qg}. We use both the paintings themselves and the comments, provided in the SemArt dataset, as the input source for generating QA pairs. In this way, the paintings provide information to generate visual questions (\eg ``What animal is this?'' in Figure \ref{fig:example_qa_pairs}), and the comments are used to generate knowledge-based questions about art (\eg ``Who depicts Napoleon in 1814?'' in Figure \ref{fig:example_qa_pairs}). 

To address this new QA task, we propose a baseline model specially designed to leverage knowledge information about art, which comes with modality selection on top of recent VQA and text QA models, which is coined as \mname{} (\textbf{vi}sual- and \textbf{k}nowkedge-branch network for predict\textbf{ing} answers). This network handles the dual-modality (visual and external knowledge-based) of the art questions with dedicated branches.

Our contributions can be summarized as follows:
\begin{itemize}
    \item[$\bullet$] Firstly, we propose a new task of art question answering, which inherently involves visual understanding of and knowledge on paintings. The latter may be deemed as textual understanding, as such knowledge can be found in books and online documents, \eg \emph{Wikipedia}. Answering questions that require both visual and textual modalities has not been well explored so far.
    \item[$\bullet$] Secondly, we build a preliminary dataset, AQUA, and make it publicly available.\footnote{\url{https://github.com/noagarcia/ArtVQA}} The QA pairs are manually cleansed by crowdsourcing workers with respect to each question's answerability and grammatical correctness, as well as the answer's correctness. 
    \item[$\bullet$] Thirdly, we present a baseline model, named \mname{}, for our art QA task. In addition to the question, the baseline model uses the painting and a paragraph retrieved from a knowledge base to predict the answer that is relevant to both the question and the painting.
    \item[$\bullet$] Finally, through the results of this study, we can envisage the challenges and possible solutions that future research aiming to address visual question answering on art must consider.
\end{itemize}
 
\section{Related Work}
\subsection{Computer Vision for Art}
Arts and computer vision have an inevitable link as many artworks have some visual components. One fundamental direction is the digitization of artworks for archiving and restoration (\eg \cite{ikeuchi2007buddha}). Several studies have been done for artworks in the computer vision field, including author/style identification \cite{shamir2010impressionism,johnson2008image,shamir2010impressionism}, image classification \cite{ma2017part,carneiro2012artistic,Tan2016CeciNP,ma2017part,garcia2019context,Huckle2020Demographic}, and image retrieval \cite{carneiro2012artistic,carneiro2012artistic,crowley2014state,crowley2015face}. To the best of our knowledge, this is the first work for question answering on paintings.

\subsection{Question Generation}
\paragraph{Visual Question Generation}
Visual question generation (VQG) can be categorized into grounded and open-ended \cite{DBLP:journals/corr/abs-1905-08949}. Grounded VQG generates questions whose answers can be found in the information relevant to the input image \cite{Zhang2017AutomaticGO}. To this end, captions are first generated from the image, and either rule-based \cite{Ren2015ExploringMA,Zhu2016Visual7WGQ} or neural \cite{Zhang2017AutomaticGO} models are used to further generate questions from the captions. Open-ended VQG are often about abstract concepts such as events and states, which can be inferred by the objects in an image \cite{mostafazadeh-etal-2016-generating}.  Diversity is crucial for open-ended VQG, for which variational auto-encoders \cite{Jain_2017_CVPR} and generative adversarial network \cite{fan-etal-2018-reinforcement} have been used.

\paragraph{Textual Question Generation} Either rule-based or neural model-based approach has been applied for textual question generation (TQG). Rule-based TQG first constructs question templates either manually \cite{heilman-smith-2010-good,mazidi-nielsen-2014-linguistic} 
or via crowdsourcing \cite{labutov-etal-2015-deep}, and then applies the templates to generate questions. \cite{du-etal-2017-learning} pioneers the first neural model for TQG, which apply the sequence-to-sequence model with attention. Neural TQG studies focus on how to encode answers \cite{zhao-etal-2018-paragraph,Kim_2019}, generate question words \cite{duan-etal-2017-question,sun-etal-2018-answer}, and use paragraph-level context \cite{zhao-etal-2018-paragraph,du-cardie-2018-harvesting}.

\subsection{Visual Question Answering}
Previous VQA studies are on either natural images or videos. Commonly used techniques for image-based VQA include joint visual and language embeddings, and attention mechanisms that model where to look in an image \cite{Wu:2017:CVIU}.
One extension of VQA is to answer questions on video. Because of the temporal information in videos, action recognition  \cite{maharaj2017dataset,jang2017tgif,zellers2018vcr,mun2017marioqa}, story understanding \cite{tapaswi2016movieqa,kim2017deepstory}, 
and temporal coherence \cite{zhu2017uncovering} have been further incorporated. Another interesting extension is to use external knowledge beyond images and videos. The knowledge can be either general \cite{wu2016ask,wang2018fvqa,marino2019ok} or dataset specific \cite{wang2017explicit,garcia2020knowit}. 

Because the acquisition of data is not always a trivial task, synthetic datasets have been commonly used by the VQA community. For example, Malinowski and Fritz \cite{malinowski2014multi} used automatically generated QA pairs based on some templates. Johnson \etal \cite{johnson2017clevr} also employed generated QA pairs on synthetic images, mainly for excluding possible biases in standard datasets. Similarly, our AQUA dataset is also synthetic and, as with the datasets in \cite{malinowski2014multi,johnson2017clevr}, we aim that it serves as a proof-of-concept for VQA on the domain of art. 

\section{AQUA Dataset}
We use the SemArt dataset \cite{Garcia2018How}, which is originally designed for semantic art understanding, as our source for generating our QA pairs. The SemArt dataset contains paintings and associated comments, where comments are blocks of text, sometimes including the metadata about the painting, such as the author name and created year. They can also have several sentences about the story in the painting and the contextual background when it was created, such as the social and the author's personal situations. These comments serve as knowledge. In order to show the potentials of AI technologies to comprehend paintings, it is important to explore techniques that work not only on the visual content in the painting themselves but also on their surrounding ideas. We therefore generate QA pairs from visual and knowledge modalities with respective question generation methods.

\subsection{Question Generation}
\paragraph{Visual Question Generation}
The inherent necessity of visual understanding makes question generation from the image content a tough problem. A number of methods have been proposed so far \cite{DBLP:journals/corr/abs-1905-08949}. We try two of them to generate a diverse set of visual questions. The first one is iQAN \cite{li2018visual} trained on VQA v2.0 \cite{goyal2017making},\footnote{The code is reproduced by ourselves, and we confirmed a similar performance to that of the original paper.} which takes an image and an answer word as input and generates a question using a neural network model. We use the object detector provided in Amazon Rekognition\footnote{\url{https://aws.amazon.com/rekognition/}} to obtain the answer words. The other one uses Pythia\footnote{\url{https://github.com/facebookresearch/pythia}} to generated a caption for each painting and transforms each generated caption into a QA pair by applying the rule-based TQG technique described below.

\paragraph{Knowledge-based Question Generation (KQG)}
For generating questions that involves the knowledge about art, we apply TQG methods, which have been studied by the natural language processing community for the last decades, relying on the natural language knowledge comments available in the SemArt dataset. We tried several TQG approaches, \ie, rule-based and neural ones. The rule-based approach \cite{heilman-smith-2010-good} builds a parsing tree from an input sentence and transforms it to QA pairs based on a set of rules. The resulting QA pairs may be filtered by statistical ranking to drop less-likely samples. The neural approach \cite{du-etal-2017-learning} is based on sequence-to-sequence modeling. We found that the rule-based technique yielded more satisfactory QA pairs.

\subsection{QA Pair Evaluation and Cleansing}

\subsubsection{Question Generation Evaluation}
\label{sec:qg_eval}
We use Amazon Mechanical Turk\footnote{\url{http://www.mturk.com}} (AMT) to evaluate the quality of our QA pairs, given a painting as well as its associated comment and question, with the following criteria:

\begin{itemize}
\item[$\bullet$] \textbf{Grammatical correctness} measures whether the QA pair is syntactically well-formed, specified by (i) \textit{no grammatical error}, (ii) \textit{minor errors} (there are some errors but the QA pair still makes sense), and (iii) \textit{major errors} (the QA pair does not make any sense).

\item[$\bullet$] \noindent\textbf{Answer existence} identifies whether the question has a clear answer in the given painting and comment.

\item[$\bullet$] \textbf{Answer correctness} measures given the QA pair whether the answer to the question is correct.

\item[$\bullet$] \textbf{Necessity of visual information} evaluates whether the visual information in the painting is needed to answer the question.

\item[$\bullet$] \textbf{Necessity of textual information} evaluates if the textual information in the comment is needed to answer the question.

\item[$\bullet$] \textbf{Question reasonability} judges whether the QA pair looks like human-generated.
\end{itemize}

\begin{table}[t!]
    \setlength{\tabcolsep}{10pt}
    \centering
    \caption{Evaluation for question generation by AMT. For grammatical correctness, the proportion of QA pairs with (i) no error and (ii) minor errors are shown, where 0.429 and 0.687 of QA pairs are with no error for VQG and KQG, respectively.}
    \label{tab:dataset_eval}
    \begin{tabular}{lrr}\toprule
    Criterion & VQG & KQG \\ \midrule
    Grammatical correctness          & 0.936 & 0.871  \\
    Answer existence                 & 0.504 & 0.842 \\
    Answer correctness               & 0.337 & 0.735 \\
    Necessity of visual information  & 0.977 & 0.514 \\
    Necessity of knowledge           & 0.098 & 0.935 \\
    Question reasonability           & 0.691 & 0.690 \\
    \bottomrule
    \end{tabular}
\end{table}

We randomly selected 1,000 and 989 QA pairs from both VQG and KQG, respectively, and evaluated them (Table \ref{tab:dataset_eval}). 
Our VGQ samples have a high grammatical correctness. However, the answer existence and correctness are low. The errors mainly come from two factors: object detection and visual encoding. Our iQAN-based VQG uses an object as input. If the object detector fails, the answer will be incorrect, which will also affect the question generation. As Amazon Rekognition is trained on real-world photos, it sometimes predicts the objects incorrectly in paintings. For the same reason, the visual encoding in iQAN and the image captioning are not as accurate as that for real-world photo datasets. This explains why many questions do not have answers in the associated painting and comment. The necessity of knowledge is low because our models tend to ask relatively simple visual questions. Yet nearly 70\% of QA pairs look like human generated.

The result for KQG shows that our generated samples also have a high grammatical quality. A common source of negative responses is pronouns in generated answers (\eg, \textit{it} and \textit{they}) because our rule-based model does not exclude pronouns in grammar trees from the candidate answer list. For 84\% of QA pairs, their answers are found in the context, and 74\% of answers are correct; a possible reason for these superior results is that the question and answer are generated together from the same grammar tree. Knowledge is required in over 93\% of QA pairs as expected because the questions are coming from the comments. Interestingly, crowd workers tend to find visual information is still necessary even for knowledge QA pairs. The question reasonability criterion shows that most QA pairs are likely to be generated by humans.

\subsubsection{Dataset Cleansing and Statistics}

\begin{table}[t]
  \centering
  \caption{Statistics on the AQUA dataset.}
  \setlength{\tabcolsep}{10pt}
  
  \label{tab:stats}
  \begin{tabular}{l r r r }
     \toprule
       & Train & Val & Test \\
      \midrule
    \# QA pairs & 69,812 & 5,124 & 4,912 \\
    \verb+  + \# Visual QA pairs  & 29,568 & 1,507 & 1,270 \\
    \verb+  + \# Knowledge QA pairs & 40,244 & 3,617 & 3,642\\
    Question length (in word) & 8.82 & 9.21 & 9.41\\
    \verb+  + for visual QA & 6.53 & 6.50 & 6.51\\
    \verb+  + for knowledge QA & 10.50 & 10.33 & 10.43\\
    Answer length & 3.13 & 3.68 & 3.85 \\
    \verb+  + for visual QA & 1.00 & 1.00 & 1.00\\ 
    \verb+  + for knowledge QA & 4.69 & 4.79 & 4.85\\
      \bottomrule
  \end{tabular}

\end{table}

To exclude QA pairs with major grammatical errors or without (correct) answers, we again use AMT. Unlike the evaluation, this time, we only evaluate grammatical correctness as well as answer existence/correctness but on the entire dataset. Table \ref{tab:stats} shows the statistics of our AQUA dataset after cleansing. Due to low answer existence/correctness, the number of visual QA pairs is smaller than that of knowledge QA pairs. The question length comes with an obvious bias because of the difference in the question generation methods. 

\subsection{Task Definition}

With our AQUA dataset, there can be several possible task definitions. In this paper, we focus on the one in which all the comments that are associated with paintings are available. More specifically, let $C = \{c_i|i=1,\dots,N\}$ denote the set of all the comments. The aim of AQUA task is to answer question $q$ given painting $v$ using $C$, without an explicit association between $v$ and a specific comment in $C$. In this task, $C$ can be viewed as an external source of knowledge, containing the necessary information to answer the question when the comment associated with $q$ is correctly retrieved. 

A more challenging extension of this task is to not use $C$ but other sources of knowledge, \eg \textit{Wikipedia}. With this extension, the performance also depends on the quality of the sources and their affinity to the original source. We leave the extension as future work.

\section{\mname{} Model}

\begin{figure}[t]
    \centering
    \includegraphics[width=\textwidth]{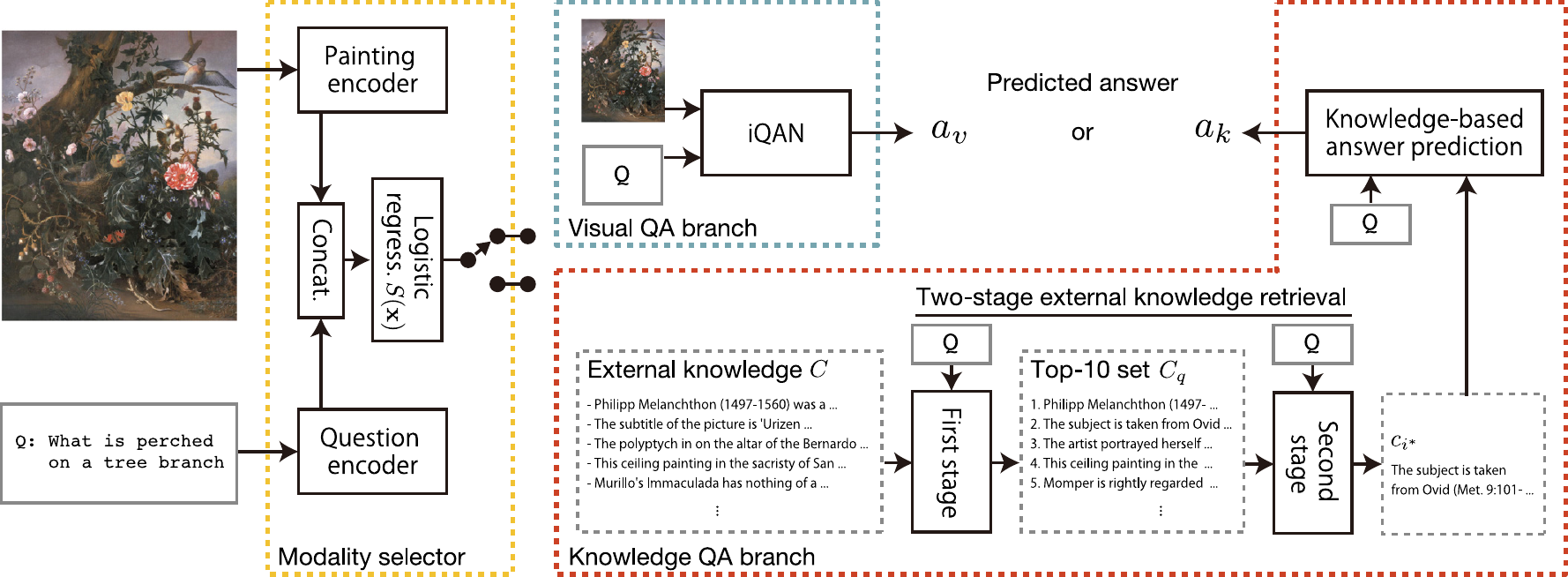}
    \caption{An overview of our \mname{} model for the AQUA dataset.}
    \label{fig:pipeline}
\end{figure}

By construction, the AQUA dataset contains two types of questions. We design our baseline model, coined \mname{}, to handle them with dedicated branches. Figure \ref{fig:pipeline} illustrates the overall pipeline. Inspired by the intuition that humans first look into the given question and then try to locate the required information to answer it (in our case, either the associated painting or comment), \mname{} consists of three main components: The question and painting are first fed into a \textit{modality selector}, which classifies the question into visual or knowledge-based ones. Questions about the visual content go through the \textit{visual QA branch}. Otherwise, questions are passed to the \textit{knowledge QA branch}, in which an associated comment is retrieved from $C$. We detail these three components below.

\subsection{Modality Selector}

Our modality selector $S$ classifies a question $q$ into these two modalities given $q$ and $v$, so that it can go through the corresponding branch. We use pretrained BERT \cite{devlin-etal-2019-bert} as the question encoder. More specifically, question $q$ is encoded into a 1,024 dimensional vector $\mathbf{q}$, which is BERT's output corresponding to the special token \texttt{[CLS]}. For our painting encoder, we use pretrained ResNet-152 \cite{he2016deep} to encode the painting into a 2,048 dimensional vector $\mathbf{v}$. We concatenate $\mathbf{q}$ and $\mathbf{v}$ into a vector $\mathbf{x}$ and train a logistic regression model %
\begin{equation*}
S(\mathbf{x}) = \frac{1}{1 + e^{-(\mathbf{w}_{\text{s}}^\top \mathbf{x} - b_\text{s})}},
\end{equation*} %
where $\mathbf{w}_\text{s}$ and $b_\text{s}$ are a trainable vector and scalar. Question $q$ is passed to the visual QA branch when $S(\mathbf{x}) > 0.5$ and to the knowledge QA branch otherwise. 

\subsection{Visual QA Branch}

Visual questions can be answered solely based on the associated painting without any external knowledge. For this question type, the task is reduced to VQA over paintings.

We again use iQAN \cite{li2018visual} as our visual QA branch, which is a dual model and can take either a question or an answer as input, then output the counterpart. We separately train the iQAN model over the training split of our AQUA dataset (\ie, we do not use the iQAN model trained for question generation). This branch produces a predicted answer $a_v$, which is from the answer vocabulary $A$ consisting of the top-5,000 most common words in the training split.

\subsection{Knowledge QA Branch}

Questions classified as knowledge-based are fed to the knowledge QA branch. We first retrieve the comment in $C$ that is the most relevant to $q$ with a two-stage strategy. In the first stage, we apply TF-IDF to rank all the comments in $C$ with respect to their relevance to $q$ and obtain the subset $C_q$ consisting of the top-10 most relevant comments. In the second stage, comments in $C_q$ are re-ranked using BERT to find the most relevant comment $c_q$. This two-stage strategy drastically reduces the computational cost compared to using BERT-based ranking directly over $C$. Finally, the answer is predicted based on the question $q$ and the retrieved comment $c_q$ with a XLNet-based model.

\subsubsection{Two-Stage External Knowledge Retrieval}
\label{sec:knowledge}

Finding the relevant comment $c_q$ is critical for this task since it contains the answer. A naive approach is to train a ranking network and apply it to all comments in $C$. This approach can be computationally expensive when $C$ contains a large number of comments and an expensive model, such as Transformer-based ones \cite{devlin-etal-2019-bert}, is used. We thus take a two-stage approach for retrieving knowledge.

For the first stage, we adopt TF-IDF to encode both $q$ and all $c_i \in C$. Letting $\hat{\mathbf{q}}$ and $\hat{\mathbf{c}}_i$ be the respective TF-IDF vectors, we compute the score $s_i = \hat{\mathbf{q}}^\top \hat{\mathbf{c}}_i / (\|\hat{\mathbf{q}}\| \, \|\hat{\mathbf{c}}_i\|)$. The set $C_q$ consists of $c_i$'s that have the 10 highest $s_i$'s. In order to improve the ranking accuracy, we apply to both $q$ and $c_i$ (i) preprocessing with NLTK\footnote{https://www.nltk.org/} for stop word removal and word stemming and (ii) $n$-gram TF-IDF where $n = 3$. 

The second stage further ranks $c_i \in C_q$ to find the comment associated with the question. We cast this into a sentence pair classification problem and use a BERT-based model to predict how likely a given $c_i$ is relevant to $q$. We concatenate $q$ and each $c_i \in C_q$ with \texttt{[SEP]} and feed it to the pretrained BERT. The output ${o}_i$ associated with \texttt{[CLS]} is passed to a logistic regression model, given by
\begin{equation*}
    R(\mathbf{o}_i) = \frac{1}{1 + e^{-(\mathbf{w}_{\text{r}}^\top \mathbf{o}_i - b_\text{r})}},
\end{equation*}
where $\mathbf{w}_{\text{r}}$ and $b_\text{r}$ are trainable vector and scalar. The model is trained as a binary classifier that predicts whether $q$ and $c_i$ are relevant or not, but its output $R(\mathbf{o}_i)$ is treated as the score for $c_i$ when inference. We use $c_{i^*}$ where $i^* = \arg \max_i R(\mathbf{o}_i)$ for answering the question. 

\subsubsection{Knowledge-Based Answer Prediction}

We use XLNet\footnote{We used XLNet instead of BERT as XLNet shows better performance on the popular Stanford question answering dataset (SQuAD2.0).} \cite{yang2019xlnet} for predicting the answer in knowledge-based questions. We concatenate the question $q$ and the $c_{i^*}$ with \texttt{[SEP]}, and feed it to XLNet, which predicts the positions of the answer starting and ending in $c_{i^*}$. We extract the words between the predicted starting and ending position as answer $a_k$. We use a pre-trained XLNet and fine-tune it over the knowledge QA pairs in our AQUA dataset. 

\section{Evaluation}
In this section, we show the performance of \mname{} as well as several more basic baselines and state-of-the-art VQA methods on the AQUA dataset.

\subsection{Evaluation Details}
The performance of our task is measured by exact match (EM), \ie the percentage of predictions that match the ground truth answer exactly. This EM-based evaluation enables us to compare baselines, \mname{}, and its variants in a single framework. It should be noted that, in the visual QA branch, the answer is the most probable word among the answer vocabulary $A$ (the top 5,000 most common answers in the training split). The upper bound of the accuracy is 0.306 if all QA pairs in the test split would went through the visual QA branch because only 1,505 QA pairs out of 4,912 have the answer in $A$. In the models that exploit external knowledge, the answer is extracted from the text in $C$.

\subsection{Baselines and \mname{} Variants}

Our first set of baselines are both blind and ignorant, answering questions without paintings and external knowledge. 

\begin{itemize}
\item[$\bullet$] \textbf{LSTM} Each word in a question is converted into word embeddings, which are trained from scratch. The word embeddings are input into a 2-layer LSTM. The hidden state of the last layer is fed into a fully connected layer classifier with the softmax activation over the answer vocabulary $A$. 

\item[$\bullet$] \textbf{BERT} Each question is input into a fine-tuned base and uncased BERT model. The special tokens \texttt{[CLS]} and \texttt{[SEP]} are added at the beginning and at the end of each sentence, respectively. The output from the first token is fed into a fully connected layer classifier followed by softmax to predict the most probable answer in the same way as the LSTM baseline.

\item[$\bullet$] \textbf{XLNet} Instead of the BERT model, XLNet is used to encode questions. The classification is done in the same way as the BERT baseline.
\end{itemize}

The second set of baselines use paintings but not external knowledge to answer questions.

\begin{itemize}
    \item[$\bullet$] \textbf{BUTD} Bottom-up and top-down attention \cite{anderson2018bottom} consists of a bottom-up module that generates object proposals from the image, and a top-down module that predicts an attention distribution over those proposals based on the question, encoded with a GRU.
    % \cite{cho-etal-2014-learning}. 
    The answers are chosen from $A$.

    \item[$\bullet$] \textbf{BAN} Bilinear attention network \cite{kim2018bilinear} also extracts a set of region proposals from the image and encodes questions with a GRU. Differently from BUTD, BAN computes attention between all the image proposals and all the words in the question.
\end{itemize}

For our \mname{} model, we have three variants, \ie, \mname{} without the knowledge QA branch (\textit{w/o K}), without the visual QA branch (\textit{w/o P}), and the \textit{full} model. In addition to them, we also evaluate the upper bound performance when the ground truth modality labels are used instead of the modality selector (\mname{} \textit{w/ L}).

\begin{table}[t]
\caption{Accuracy for different methods on the AQUA test split. Q, P, K, and L stand for questions, paintings, external knowledge, and labels, which are the information used in the respective models.}
\label{tab:results}
\setlength{\tabcolsep}{7pt}
\centering
\begin{tabular}{l c c c c c }
\toprule
Method & Q & P & K & L & EM\\
\midrule
LSTM & \cmark & - & - & - & 0.198 \\
BERT & \cmark & - & - & - & 0.194 \\
XLNet & \cmark & - & - & - & 0.193 \\
\midrule
BUTD & \cmark & \cmark & - & - & 0.218 \\
BAN & \cmark & \cmark & - & - & 0.224 \\
\midrule
\mname{} \textit{w/o K} & \cmark & \cmark & - & - & 0.204 \\
\mname{} \textit{w/o P} & \cmark & - & \cmark & - & 0.352 \\
\mname{} \textit{full} & \cmark & \cmark & \cmark & - & 0.555 \\
\midrule
\mname{} \textit{w/ L} & \cmark & \cmark & \cmark & \cmark & 0.555 \\
\bottomrule
\end{tabular}

\end{table}

\subsection{Results Analysis}

\paragraph{Model Comparison} Results are presented in Table \ref{tab:results}.
As expected, methods relying only on questions to answer perform poorly, showing that the task requires the information from multiple inputs for answering correctly. When the paintings are added into the system, as in BUTD and BAN, performance improves up to 0.224. However, they lag well behind the accuracy obtained with our proposed \mname{} that leverages the information from external sources of information. Overall, our proposed model outperforms other methods, including BUTD and BAN, by a huge margin. 

\paragraph{\mname{} Variants} Our full model improves by more than 0.351 and 0.203 compared to the results of the visual and the knowledge QA branch only models, respectively, showing the benefits of using both the visual information obtained from the paintings and the information obtained from external knowledge. Also, we note that the use of ground truth labels instead of the modality selector hardly affects the overall performance. This implies the modality selector's efficiency.

\begin{figure}[t!]
    \centering
    \includegraphics[width=0.95\columnwidth]{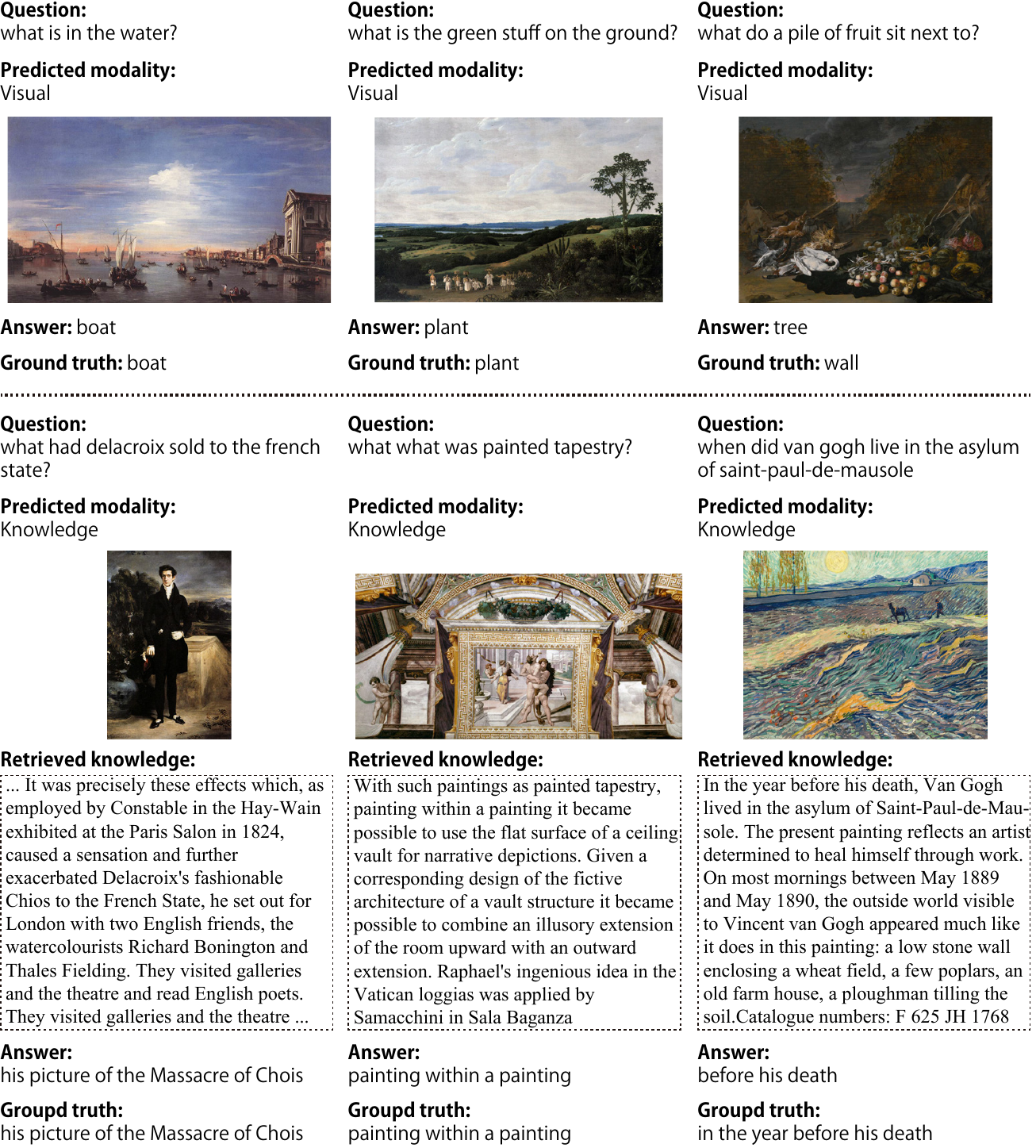}
    \caption{VIKING \textit{full} results for visual (up) and knowledge (down) questions. The right-most column shows incorrect predictions for both modalities.}
    \label{fig:examples_result}
\end{figure}

\begin{figure}[t!]
    \centering
    \includegraphics[width=0.9\columnwidth]{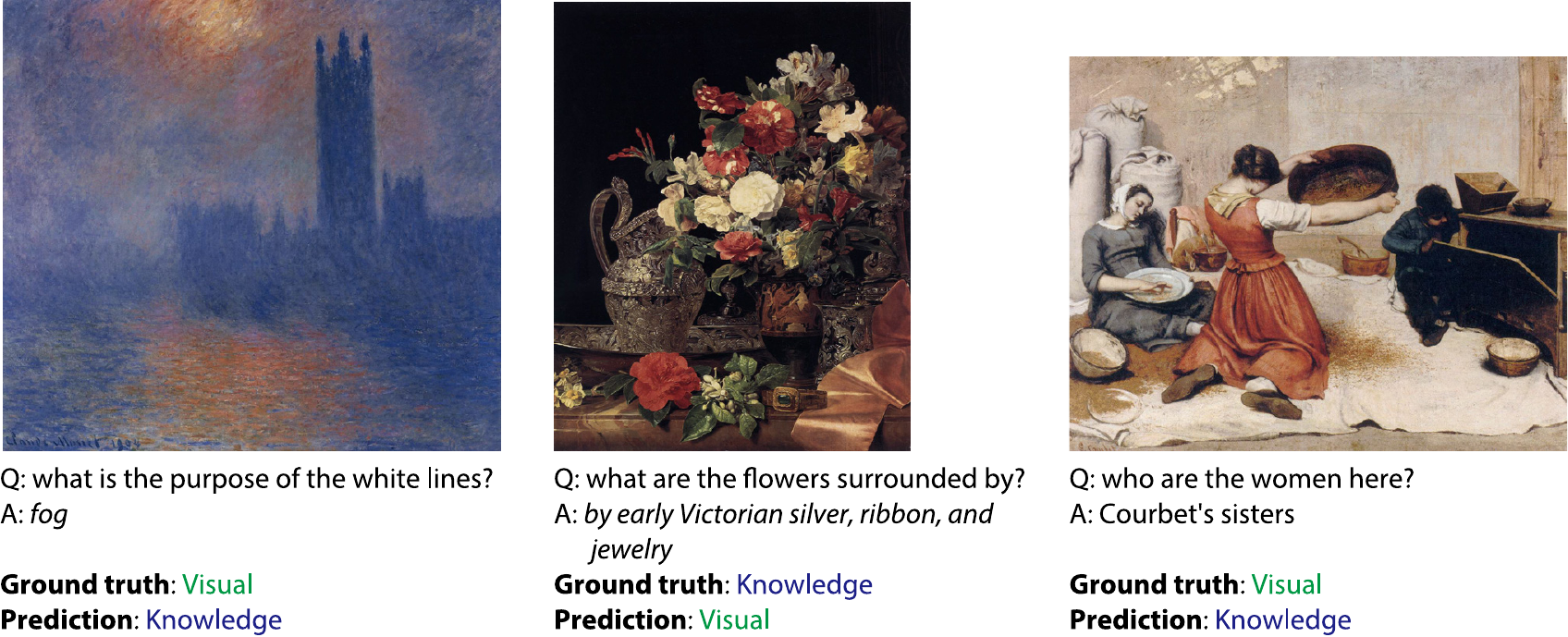}
    \caption{Failures in modality selector.}
    \label{fig:ms_failure}
\end{figure}

\paragraph{Qualitative Results} Figure \ref{fig:examples_result} shows example predictions by \mname{} \textit{full}. The modalities of all six examples are correctly predicted. The dataset sometimes contains question and answer pairs that are not obvious. The top-right example illustrates this problem, in which it is not clear if there is a wall or any other things next to the fruit. For the bottom-right example, the ground-truth answer is ``in the year before his death,'' while \mname{} predicted ``before his death.'' Semantically, the prediction is almost correct, but it is counted as incorrect due to the EM-based evaluation.

\subsection{Evaluation of Knowledge-Related Components}
Next, we study the performance of the components involving external knowledge.

\paragraph{Modality Selector}

\begin{table}[t]
\caption{Confusion matrix of the modality selector.} 
\setlength{\tabcolsep}{7pt}
  \centering
  \begin{tabular}{l r r }
     \toprule
      & \multicolumn{2}{c}{Prediction} \\
      \cline{2-3}
      Label & Visual & Knowledge \\
      \midrule
      Visual & 1,269 & 1 \\
      Knowledge & 17 & 3,625 \\
     \bottomrule
  \end{tabular}
  \label{tab:confusionmatrix}
\end{table}

Table \ref{tab:confusionmatrix} shows the confusion matrix of the modality selector on the test split. Since the visual and the knowledge questions are generated using different methods, it is relatively easy for the classifier to distinguish them, getting an accuracy of 0.996. This result supports the fact that there is no gain between \mname{} \textit{full} and \textit{w/ L}. Most failures in the modality selector (Figure \ref{fig:ms_failure} gives some examples) are reasonable, asking questions that appear to require the other modality.  

\paragraph{External Knowledge Retrieval}
The performance of the external knowledge retrieval is reported in Table \ref{tab:retrieval}, together with its variants. The performance is measured as recall at $k$ (R@$k$), \ie the percentage of QA pairs whose original comment is ranked in the top $k$ positions. Our two-stage external knowledge retrieval achieves the highest performance. Specifically, the full variant (\ie, TF-IDF + PP + $n$-gram, where PP stands for preprocessing) of the first stage ranked the original comments within top-10 for over 90\% of QA pairs, whereas the second stage gains over 5\% by the BERT-based ranking.

\begin{table}[t]
\caption{External knowledge retrieval performance. PP stands for preprocessing.} \label{tab:retrieval}
\setlength{\tabcolsep}{7pt}
  \centering
  \begin{tabular}{l c c c c }
     \toprule
      First stage & Second stage & R@1 & R@5 & R@10 \\
      \midrule
      TF-IDF & - & 0.588 & 0.775 & 0.822  \\
      TF-IDF + PP & - & 0.600 & 0.803 & 0.844 \\
      TF-IDF + PP + $n$-gram & - & 0.712 & 0.878 & 0.907 \\
      TF-IDF + PP + $n$-gram & \cmark & 0.769 & 0.879 & 0.907 \\
      \bottomrule
  \end{tabular}%}
  
\end{table}

\section{Discussion}
This work is presented as a concise first approximation to the task of art-based visual question answering and it aims to set the foundations for incorporating art knowledge in a computer vision system. However, despite the encouraging results obtained in our experimental evaluation, it presents some limitations.

\paragraph{Dataset Limitations:} The questions and answers in our proposed AQUA dataset are automatically generated from paintings and their associated comments. This process presents some limitations: 1) questions and their answers are either relate to the visual content or to the background knowledge, but usually not both. It would be interesting to incorporate samples where both are needed, increasing the complexity of the problem; 2) the visual questions are based on the labels extracted by an object detector trained on real-world photos, which introduces noise specially on paintings with non-realistic styles; and 3) because of the automatic generation of answers,  the variety of the questions and of the capabilities required to answer them is rather limited, \eg, the answers of visual questions can only be detected objects in the images.

\paragraph{Baseline Limitation:} We introduced VIKING as a baseline model for art-based VQA. VIKING is built on top of state-of-the-art VQA and TQA models. Apart from the specific limitations of those systems, VIKING presents two domain specific limitations: 1) on the visual part, the model is applied equally to all the painting images, without considering the differences on artistic styles. Incorporating style correction techniques would benefit the visual recognition part, specially for object detection; 2) on the knowledge part, VIKING uses the same source of information as in the question generation process (\ie, comments from the SemArt dataset). A more realistic setting would require to query independent sources of external knowledge, such as Wikipedia. 

Considering these limitations, we envisage some promising future directions on art-based VQA. On the dataset construction part, it would be interesting to incorporate human expert question-answer pairs that require both visual and knowledge understanding. This would increase the complexity and relevance of the dataset. On the model design part, the addition of specific features related to paintings, differing from the real-world images, would probably improve the model performance. For example, adaptation between the real-world and the painting domain in the object detector, or disentanglement of content and style.

\section{Conclusion}
This paper proposes a new task of visual question answering on art pieces, and presents our preliminary dataset on this task, coined AQUA, which consists on automatically generated visual and knowledge-based QA pairs. We also present a model called \mname{} that serves as a baseline for future exploration on this task. \mname{} leverages complementary information in paintings and external knowledge with a two branch model. Our experimental results demonstrated that \mname{} outperformed existing models with a large margin, which means that the model is a strong baseline to compare. The task definition in this paper assumes that the external knowledge is strongly tied with the questions (\ie, the comments used for generating the QA pairs are available for the QA module). This may be in a sense viewed as the upper bound of the performance. Using other sources of external knowledge will be our future direction.

\textbf{Acknowledgment} This work was partly supported by JSPS KAKENHI Nos. 18H03264 and 20K19822, and JST ACT-I. 

%\clearpage
\bibliographystyle{splncs04}
\bibliography{ijcai20,semart,icmr2019,aaai,egbib}
\end{document}